 \documentclass[final,5p,times,twocolumn]{elsarticle}

\usepackage{cite}
\usepackage{amsmath}
\usepackage{amssymb}
\usepackage{graphicx}
\usepackage{algorithm}
\usepackage{algorithmic}
\usepackage{tikz}
\usepackage{tikz-qtree}
\usepackage{multirow}
\usepackage{rotating}
\usepackage{subfigure}
\usepackage{framed}
\usepackage{color}
\usepackage{colortbl}
\usepackage{longtable}
\usepackage{supertabular}
\usepackage{booktabs}
\usepackage{url}
\definecolor{shadecolor}{rgb}{1,0,0}

\begin{document}

\begin{frontmatter}



\title{A Many-Objective Evolutionary Algorithm Based on  Decomposition and Local Dominance}


\author[lcu]{Yingyu Zhang}
\ead{zhangyingyu@lcu-cs.com}
\author[lcu]{Yuanzhen Li}
\ead{liyuanzhen@lcu-cs.com}
\author[su]{Quan-Ke Pan\corref{cor}}
\ead{panquanke@shu.edu.cn}
\author[ntu]{P.N. Suganthan}
\ead{epnsugan@ntu.edu.sg}
\cortext[cor]{Corresponding author.}
\address[lcu]{School of Computer Science, Liaocheng University, Liaocheng 252059, China}
\address[su]{School of Software Engineering, South China University of Technology, Guangzhou 510006, China}
\address[ntu]{School of Electrical and Electronic Engineering, Nanyang Technological University, Singapore 639798, Singapore}

\begin{abstract}
Many-objective evolutionary algorithms (MOEAs), especially the decomposition-based MOEAs, have attracted wide attention in recent years. Recent studies show that a well designed combination of the decomposition method and the domination method can improve the performance ,i.e., convergence and diversity, of an MOEA. In this paper, a novel way of combining the decomposition method and the domination method is proposed. More precisely, a set of weight vectors is employed to decompose a given many-objective optimization problem (MaOP), and a hybrid method of the penalty-based boundary intersection function and dominance is proposed to compare local solutions within a subpopulation defined by a weight vector.
The MOEA based on the hybrid method is implemented and tested on problems chosen from two famous test suites, i.e., DTLZ and WFG. The experimental results show that our algorithm is very competitive in dealing with MaOPs.
Subsequently, our algorithm is extended to solve constraint MaOPs, and the constrained version of our algorithm also shows good performance in terms of convergence and diversity.
These reveals that using dominance locally and combining it with the decomposition method can effectively improve the performance of an MOEA.
\end{abstract}


\end{frontmatter}

\section{Introduction}
A lot of  real-world problems such as electric power system reconfiguration problems \citep{Panda2009937},
water distribution system design or rehabilitation problems \citep{WBS2013}, automotive engine calibration problems \citep{AEC2013}, land use management problems\citep{LUM2012}, optimal design problems \citep{Ganesan2015293,Ganesan2013,DomingoPerez201695}, flowshop scheduling problems \citep{Pan2011, Pan2018, Pan2019},
and problems of balancing between performance and cost in energy systems \citep{Najafi201446}, etc.,
can be formulated into multi-objective optimization problems (MOPs) involving more than one objective function.
Without loss of generality, an MOP can be considered as a minimization problem as follows:
\begin{equation}\label{MOP}
\begin{split}
Minimize \quad &F(x)=(f_1(x),f_2(x),...,f_M(x))^T \\
 &Subject \quad to \quad x\in\Omega,
 \end{split}
\end{equation}
where $M\geq 2$ is the number of objective functions, $x$ is a decision vector,  $\Omega$ is the feasible set of decision vectors,
and $F(x)$ is composed of M conflicting objective functions. An MOP is usually referred to as a many-objective optimization problem (MaOP) when M is greater than 3.

A solution $x$ of Eq.(\ref{MOP}) is said to dominate the other one $y$ ($x\preccurlyeq y$),
if and only if $f_i(x)\leq f_i(y)$ for $i\in(1,...,M)$ and $f_j(x)<f_j(y)$ for at least one index $j\in(1,...,M)$.
It is clear that $x$ and $y$ are non-dominated with each other, when both $x\preccurlyeq y$ and $y \preccurlyeq x $ are not satisfied.
A solution $x$ is Pareto-optimal to Eq.(\ref{MOP}) if there is no solution $y\in\Omega$  such
that $y\preccurlyeq x$. $F(x)$ is then called a Pareto-optimal objective vector. The
set of all the Pareto optimal objective vectors is the PF \citep{PF}.
The goal of an MOEA is to find a set of solutions, the corresponding objective vectors of which are approximate to the PF.

Multi-objective evolutionary algorithms (MOEAs) are  popular in solving MOPs, such as the non-dominated sorting genetic algorithm-II (NSGA-II) \citep{NSGAII},
the strength pareto evolutionary algorithm 2 (SPEA-2) \citep{SPEA2},
and the multi-objective evolutionary algorithm based on decomposition (MOEA/D) \citep{MOEAD}, etc.
In General, MOEAs can be divided into three categories \citep{Survey2017}.  The first category is known as the indicator-based MOEAs.
In an indication-based MOEA, the fitness of an individual is usually evaluated by a performance indicator
such as hypervolume \citep{Emmerich2005}.
Such a performance indicator is designed to measure the convergence and diversity of the MOEA,
and hence expected to drive the population of the MOEA to converge to the Pareto Front(PF) quickly with good distribution.
The second category is  the domination-based MOEAs, in which the domination method plays a key role.
However, in the domination-based MOEAs, other measures  have to be adopted  to maintain  the population diversity.
In NSGA-II, crowding distances of all the individuals are calculated at each generation and used to keep the population diversity ,
while reference points are used in NSGA-III \citep{NSGAIII}.
The third category  is the decomposition-based MOEAs.
In a decomposition based MOEA, an MOP is decomposed  into a set of subproblems and then optimized simultaneously.
A uniformly generated set of weight vectors associated with a fitness assignment method such as the weighted sum approach,
the Tchebycheff approach and the penalty-based boundary intersection (PBI) approach,  is usually used to decompose a given MOP.
Generally, a weight vector determines a subproblem  and defines a neighborhood.
Subproblems in a neighborhood are expected to own similar solutions and might be updated by a newly generated solution.
The decomposition-based MOEA framework emphasizes the convergence and diversity of the population in a simple model.
Therefore, it was studied extensively and improved from different points of view \citep{Survey2017, RVEA,MOEADD}
since it was first proposed by Zhang and Li in 2007 \citep{MOEAD}.

Although conventional MOEAs have achieved great success in solving MOPs, they often fail to solve MaOPs.
The reasons behind the failure can be attributed to the weakness of selection pressure exerted by the criterion for comparing solutions and the loss of population diversity  in the process of evolution.
The weakness of selection pressure slows down the convergence speed of an MOEA, and the loss of population diversity leads to a very poor distribution of the resulting population.
Therefore, balancing between the convergence and diversity becomes a critical issue for evolutionary algorithms to solve MaOPs.

A lot of efforts have been made to deal with this issue, and some methods are introduced into MOEAs to keep balance between convergence and diversity.
One of the most successful evolutionary algorithms for solving MaOPs may be the evolutionary many-objective optimization algorithm based on dominance and decomposition (MOEA/DD)  proposed in \citep{MOEADD}.

In MOEA/DD, each individual is associated with a subregion uniquely determined by a weight vector,
and each weight vector (or subregion) is assigned to a neighborhood.
In an iterative step, mating parents is chosen from the neighboring subregions of the current weight vector with a given probability $\delta$,
or the whole population with a low probability $1-\delta$. In case that no associated individual exists in the selected subregions,
mating parents are randomly chosen from the whole population.
And then serval classical genetic operators such as the simulated binary crossover (SBX) \citep{SBX} and the polynomial mutation \citep{PM}, etc.,
are applied on the chosen parents to generate an offspring.
Subsequently, the offspring is used to replace the worst solution within the current population
determined by a hybrid method based on decomposition and dominance.
All the solutions are arranged into N subregions, and divided into multiple levels according to their dominance relationship.
To determine the worst solution, the most crowded region is first identified from the subregions associated with the solutions in the last domination level, and the solution with the largest PBI value is selected as the worst one.

Like MOEA/D, MOEA/DD uses a set of weight vectors to decompose a given MOP into a set of subproblems and optimizes them simultaneously. However, their update strategies are very different.
In MOEA/D,  multiple solutions in current neighborhood might be replaced at the same time by the newly generated offspring. As for MOEA/DD, only the worst solution in the population is replaced by the newly generated offspring.
Besides, MOEA/DD also incorporates the domination method into its update strategy to help select the worst solution.

As it can be seen in MOEA/DD, the dominance method is applied globally on the whole population to divide the whole population into one or more domination levels.
In this paper, the domination method is only used in local subpopulation (subregion), and a so-called many-objective evolution algorithm based on decomposition and local dominance (MOEA/DLD) is proposed.
In detail, MOEA/DLD employs a set of weight vectors to decompose the current population into $N$ subpopulations, but uses dominance locally along with the PBI function to compare solutions within a subpopulation defined by a weight vector. A solution $x$ is considered to be better than the other one $y$, if $x$ dominates $y$. In the case that, none of the two dominates the other, then the one with the smaller PBI value is the better.
According to such a comparison criterion, the solutions within each subpopulation can be ordered from good to bad.
The elitism strategy is then applied to select $N$ elitist individuals to form the next generation.



The rest of the paper is organized as follows.
In Section II, the algorithm MOEA/DLD is proposed. A general framework of it is first presented.
Subsequently, the initialization procedure and the many body of the evolution are elaborated.
Some discussions about the similarities and differences between MOEA/DLD and several other algorithms are also made.
In Section III, the performance metrics,i.e., Inverted Generational Distance (IGD)  and  Hypervolume (HV) , used to evaluate the performance of the algorithm, are introduced.
In Section IV, Some features of the benchmark problems, i.e., DTLZ1 to DTLZ4 and WFG1 to WFG9,  are summarized.
In Section V, experimental results of MOEA/DLD on the benchmark problems are compared to those of other MOEAs.
In Section VI, MOEA/DLD is extended to solve constraint MOPs.
The paper is concluded in Section VII.


\section{Proposed Algorithm MOEA/DLD}

\begin{algorithm}
\caption{Main Framework of MOEA/DLD}
\label{algFramework}
\begin{algorithmic}[1]
\ENSURE  Final Population.
\STATE   Initialization.
\STATE  t=1;
\WHILE{$t<=t_{max}$}
\STATE $Q_t$=reproduction($P_t$);
\STATE $\{P_{t,1},P_{t,2},...,P_{t,N}\}$=population-partition($P_t$,$Q_t$);
\STATE $P_{t+1}$=elitist-selection($\{P_{t,1},P_{t,2},...,P_{t,N}\}$);
\STATE $t=t+1$;
\ENDWHILE
\RETURN $P_{t_{max}}$;
\end{algorithmic}
\end{algorithm}

\subsection{Initialization}
The initialization procedure includes five steps: 1) Generate uniform distributed weight vectors using a systematic sampling approach (SSA). More exactly, the original SSA \citep{SSA} is used to generate weight vectors for instances with objective number of 3 and 5. The number of the weight vectors is then calculated as
\begin{equation}\label{numberOfWeightVectors}
N(D,M)=\left(
\begin{array}{c}
  D+M-1 \\
  M-1
\end{array}
\right)
\end{equation}
where $D > 0$ is the number of divisions along each objective coordinate.
And the SSA-based two-layer weight vector generation method used in \citep{NSGAIII} is applied for instances with objective number more than 5.
At first, a set of $N_1$ weight vectors in the boundary layer and a set of $N_2$ weight vectors in the  inside layer are generated,
according to the systematic sampling approach described above.
Then, the coordinates of weight vectors in the inside layer are shrunk by a coordinate transformation as
\begin{equation}
v^{j}_{i}=\frac{1-\tau}{M}+\tau\times \omega^{j}_{i},
\end{equation}
where $\omega^{j}_{i}$ is the ith component of the jth weight vectors in the  inside layer, and $\tau\in [0,1]$ is a shrinkage factor set as
$\tau=0.5$ in \citep{NSGAIII} and \citep{MOEADD}.
At last, the two sets of weight vectors are combined to form the final set of weight vectors.
Denote the numbers of the weight vectors generated in the boundary layer and the inside layer as $D1$ and $D2$ respectively.
Then, the number of the weight vectors generated by the two-layer weight vector generation method is
$N(D1,M)+N(D2,M)$.

2) Find  neighbors for each weight vector.  To find a weight vector's neighbors, the included angles between the weight vector and all weight vectors are first calculated. And then $T$ vectors with minimum angles are selected as the neighbors of the weight vectors. The included angle between two weight vectors $w_i$ and $w_j$ can be calculated as
\begin{equation}\label{angle}
\tan\theta=\frac{d2}{d1},
\end{equation}
where
\begin{equation}\label{TwoDists}
d_1=\frac{\left\|w_i^{T}w_j\right\|}{\|w_j\|}, \quad d_2=\left\|w_i-d_1\frac{w_j}{\|w_j\|}\right\|.
\end{equation}

3) Randomly generate $N$ individuals as the initial population. 4) Find the minimum values for all the objectives to form the current ideal point. 5) Find the maximum values for all the objectives to form the current nadir point.

\subsection{Reproduction}
The reproduction procedure contains three steps: 1) mating selection, which runs over $N$ weight vectors to choose  $N$ pairs of mating parents for offspring generation. In MOEA/DLD, each weight vector is assigned with a neighborhood based on angle. For each weight vector (or neighborhood), a pair of mating parents is selected from the neighborhood with a probability $\delta$, or from the whole population with a probability $1-\delta$. 2) crossover, while generates $N$ pairs of offsprings by applying the SBX operator on the $N$ pairs of mating parents, and preserves $N$ offspring by abandoning one offspring from each pair of the offsprings. 3) mutation, which generates $N$ new offsprings constituting the new population $Q_t$ of the tth generation by applying the polynomial mutation operator on the $N$ preserved offsprings.

\subsection{Population partition}
Individuals in the old population $P_t$ and the new population $Q_t$ of the tth generation are divided into $N$ subpopulations $P_{t,1}$, $P_{t,2}$, . . . , $P_{t,N}$ by associating each individual with its closest weight vector (associated weight vector). Individuals in each subpopulation are arranged in order of good to bad. For an individual, finding its closest weight vector is to find a weight vector that has the smallest included  angle with the individual. Once the associated weight vector of an individual is found, it will be inserted into the subpopulation keeping the order of good to bad unchanged. All the individuals are divided into $N$ subpopulations, by inserting them into their associated subpopulations.

The calculation of the included angle between an individual and a weight vector $w$ is similar to that between two weight vectors, but the objective values of the individual need to be translated by subtracting the ideal point. Therefore, calculations of the two Euclidean distances become:
\begin{equation}\label{TwoDists_Norm}
\begin{split}
&d_1=\frac{\left\|(F(x)-z^{*})^{T}w\right\|}{\|w\|}\\
&d_2=\left\|F(x)-\left(z^{*}+d_1\frac{w}{\|w\|}\right)\right\|,
\end{split}
\end{equation}
where $z^{*}=(z_1^{*},z_2^{*},...,z_M^{*})^T$ is the ideal point.
And the included angle between an individual and a weight vector can be calculated by using Eq. (\ref{angle}).

The comparison operation $compare(x,y,w)$ is presented in Algorithm \ref{comparison}. To compare two individuals, a hybrid method based on dominance and the PBI function is proposed. The PBI value of an individual $x$ corresponding to a weight vector $w$ is calculated as
\begin{equation}
PBI(x|w,z^{*})=d_1+\theta d_2,
\end{equation}
where $\theta$ is the penalty factor.
The dominance between the two solutions is compared first. A solution $x$ is considered better that the other one $y$, if $x$ dominates $y$. In the case that, none of them dominates the other, their PBI values corresponding to the associated weight vector are compared. The solution with smaller PBI value is the better.

\begin{algorithm}
\caption{Comparison Procedure:$compare(x,y,w)$}
\label{comparison}
\begin{algorithmic}[1]
\REQUIRE    two individuals $x$, $y$; the associated weight vector $w$.
\ENSURE     true if $x$ is better than $y$, or false otherwise.
\IF{$x\preccurlyeq y$}
\RETURN \TRUE;
\ELSIF{$y\preccurlyeq x$}
\RETURN \FALSE;
\ELSE
\RETURN {$PBI(x|w,z^{*})<PBI(y|w,z^{*})$};
\ENDIF
\end{algorithmic}
\end{algorithm}

\subsection{Elitist selection}
As it is mentioned above, the population partition procedure divide all the individuals in the new population and old population into $N$ subpopulations, and individuals in each subpopulation are ordered from good to bad. From a different point, the first individuals from all the subpopulations form the first level, and the second individuals form the second level, and so on.

\begin{figure}[!htbp]
\centering                                                   %
\includegraphics[scale=0.65]{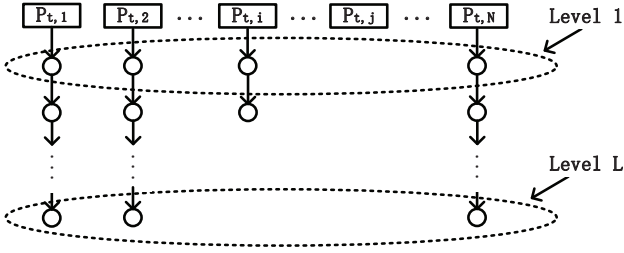}
\caption{Illustration of subpopulations and levels.}
\label{Subpopulations}                                                        
\end{figure}

As shown in Fig. \ref{Subpopulations}, all individuals can be partitioned into multiple levels.
Since a subpopulation can have zero, one or more individuals,  the number of individuals at a level is greater than or equal to that at the  next level. In addition, the higher the level, the worse the individuals.
Let $N_{t+1}$ be the number of individuals in  $P_{t+1}$, $N_i$ be the number of individuals in the ith level.
The total number of individuals is $\sum_{i=1}^{L}N_i=2N$.
The elitist selection procedure is presented in Algorithm \ref{elististSelection}.
At the beginning, $P_{t+1}$ is empty, i.e., $N_{t+1}=0$.
The main idea of the elitist selection procedure is to add individuals into $P_{t+1}$ level by level, until $N_{t+1}+N_i>N$.
Finally, choose $N-N_{t+1}$ individuals randomly from the ith level and add them into $P_{t+1}$.

\begin{algorithm}
\caption{Elitist Selection Procedure}
\label{elististSelection}
\begin{algorithmic}[1]
\REQUIRE    Subpopulations.
\ENSURE     The next generations $P_{t+1}$.
\STATE      $N_{t+1}=0$; i=1;
\WHILE{$N_{t+1}+N_i<=N$}
\STATE      Add individuals in the ith level into $P_{t+1}$;
\STATE      $N_{t+1}=N_{t+1}+N_i$;
\STATE      $i=i+1$;
\ENDWHILE
\STATE      Randomly choose $N-N_{t+1}$ individuals from the ith level and add them into $P_{t+1}$;
\RETURN     $P_{t+1}$;
\end{algorithmic}
\end{algorithm}

\subsection{Objective Normalization}
Objective normalization has been proven to be effective \citep{MOEAD, Cvetkovic2002MOP, Osiadacz1989MOP, Coit1998Genetic} for an MOEA to solve MOPs with disparately scaled objectives such as ZDT3 and WFG1 to WFG9. In this paper, we adopt a simple  normalization method \citep{MOEAD} that transforms each objective according to the following form to solve MOPs with disparately scaled objectives:
\begin{equation}
\bar{f}_i(x)=\frac{f_i(x)-z^{*}_{i}}{z_i^{nad}-z^{*}_{i}},
\end{equation}
where $z^{*}_i=\min\{f_i(x)|x\in PS\}$ and $z_i^{nad}=\max\{f_i(x)|x\in PS\}$, i.e., $z^{*}=(z_1^{*},z_2^{*},...,z_M^{*})^T$ is the ideal point, and $z^{nad}=(z_1^{nad},z_2^{nad},...,z_M^{nad})^T$ is the nadir point.
Since it is generally not easy to obtain $z^{*}$ and $z^{nad}$ in advance,
we replace $z^{*}_i$ and $z_i^{nad}$ with the minimum and maximum values of the ith objective that have been found so far, respectively.
In this way, Eq. (\ref{TwoDists_Norm}) can be rewritten as
\begin{equation}
\begin{split}
&d_1=\frac{\left\|(\bar{F}(x))^{T}w\right\|}{\|w\|}\\
&d_2=\left\|\bar{F}(x)-d_1\frac{w}{\|w\|}\right\|,
\end{split}
\end{equation}
where $\bar{F}(x)=(\bar{f}_1(x),\bar{f}_2(x),...,\bar{f}_M(x))^T$.

\subsection{Discussion}
This section highlights the similarities and differences of MOEA/DLD, MOEA/D, and MOEA/DD.
\begin{enumerate}
  \item MOEA/DLD and MOEA/DD can be seen as variants of MOEA/D to some extent, since all of the three algorithms employ a set of weight vectors to decompose a given MOPs into a set subproblems and optimize them simultaneously. In other words, the weight vectors are used to guide the evolution process in all of them.
  \item In MOEA/DLD, MOEA/D and MOEA/DD, each pair of mating parents is selected from the neighborhood with a probability $\delta$, or from the whole population with a probability $1-\delta$.
  \item The PBI function and the domination method are used to compare solutions in MOEA/DLD and MOEA/DD. In MOEA/D, the Weighted Sum Approach, the PBI approach and the Tchebycheff approach are optional for
      comparing solutions. But the PBI approach is the main choice in MOEA/DLD and MOEA/DD,
      though the Weighted Sum Approach  and  the Tchebycheff approach are also feasible in principle.
  \item In MOEA/DLD, the domination method is employed to compare local solutions within a subpopulation. However it is used as a ranking method for the whole population in MOEA/DD.
  \item Both MOEA/D and MOEA/DD are steady-state evolutionary algorithms, but MOEA/DLD is an evolutionary algorithm using elitism strategy.
\end{enumerate}

\subsection{Time Complexity}
To find  the associated weight vector of an individual, it needs to  calculate the included angle between  each weight vector and the individual, and  find the weight vector having the smallest included angle with the individual.
This takes $O(MN)$ floating-point operations. And it needs $O(1)$ comparison operations on average and $O(N)$ comparison operations in the worst case to insert the individual into the associated subpopulation.
Therefore, it takes  $O(NM)$  operations on average and $O(N^2M)$ operations in the worst case
for the population partition procedure to divide $2N$ individuals into $N$ subpopulations.
In addition, the elitist selection procedure takes  $O(N)$  operations on average
and $O(N^2)$ operations in the worst case to select $N$ elitists to form the next generation.

On the whole, the average time complexity and the worst case time complexity of each generation are
$O(N^2M)$ and $O(N^3M)$ respectively, which are the same as those of MOEA/DD, but worse than those of MOEA/D.

\section{Performance Metrics}
\subsection{Inverted Generational Distance(IGD)}
Let $S$ be a result solution set of an MOEA on a given MOP.
Let $R$ be a set of uniformly distributed representative points of the PF.
The IGD value of $S$ relative to $R$ can be calculated as \citep{IGD}
\begin{equation}
IGD(S,R)=\frac{\sum_{r\in R}d(r,S)}{|R|}
\end{equation}
where $d(r,S)$ is the minimum Euclidean distance between $r$ and the points in $S$, and $|R|$ is the cardinality of $R$. Note that, the points in $R$ should be well distributed and $|R|$ should be large enough to ensure that the points in $R$ could represent the PF very well. This guarantees that the IGD value of $S$ is able to measure the convergence and diversity of the solution set. The lower the IGD value of $S$, the better its quality \citep{MOEADD}.

\subsection{Hypervolume(HV)}
The HV value of a given solution set $S$ is defined as \citep{HV}
\begin{equation}
HV(S)=vol\left( \bigcup_{x\in S}\left[ f_1(x),z_1 \right]\times \ldots \times\left[ f_M(x),z_M \right]\right),
\end{equation}
where $vol(\cdot)$ is the Lebesgue measure,and $z^r=(z_1,\ldots,z_M)^T$ is a given reference point. As it can be seen that the HV value of S is a measure of the size of the objective space dominated by the solutions in S and bounded by $z^r$.

As with \citep{MOEADD}, an algorithm based on Monte Carlo sampling proposed in \citep{HYPE}  is applied to compute the approximate HV values for 15-objective test instances, and the WFG algorithm \citep{WFGalgorithm} is adopted to compute the exact HV values for other test instances for the convenience of comparison. In addition, all the HV values are normalized to $[0,1]$ by dividing $\prod_{i=1}^{M}z_i$.

\section{Benchmark Problems}
\subsection{DTLZ test suite}
Problems DTLZ1 to DTLZ4 from the DTLZ test suite proposed by Deb et al \citep{DTLZ}  are chosen for our experimental studies in the first place.
One can refer to \citep{DTLZ} to find their definitions.
Here, we only summarize some of their features.
\begin{itemize}
  \item DTLZ1:The global PF of DTLZ1 is the linear hyper-plane $\sum_{i=1}^{M}f_i=0.5$. And the search space contains $(11^k-1)$
  local PFs that can hinder an MOEA to converge to the hyper-plane.
  \item DTLZ2:The global PF of DTLZ2 satisfys $\sum_{i}^{M}f_i^2=1$.
  Previous studies have shown that this problem is easier to be solved by existing MOEAs,
   such as NSGA-III, MOEADD, etc., than DTLZ1, DTLZ3 and DTLZ4.
  \item DTLZ3:The definition of the glocal PF of DTLZ3 is the same as that of DTLZ2.
  It introduces $(3^k-1)$ local PFs. All local PFs are parallel to the global PF
  and an MOEA can get stuck at any of these local PFs before converging to the global PF.
  It can be used to investigate an MOEA's ability to converge to the global PF.
  \item DTLZ4:The definition of the global PF of DTLZ4 is also the same as that of DTLZ2 and DTLZ3.
  This problem can be obtained by modifying DTLZ2 with a different meta-variable mapping,
  which is expected to introduce a biased density of solutions in the search space.
  Therefore, it can be used to investigate an MOEA's ability to maintain a good distribution of solutions.
\end{itemize}

\begin{figure}[!htbp]
\begin{center}                                                       
\subfigure[DTLZ1]{                    
\includegraphics[scale=0.4]{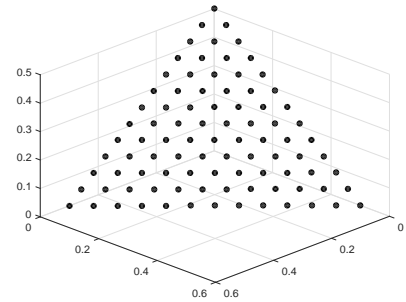}
}

\subfigure[DTLZ2-DTLZ4]{                    
\includegraphics[scale=0.4]{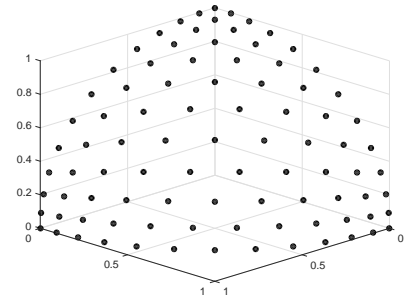}                
}

\end{center}

\caption{The representative points for the global PFs of the 3-objective instances of DTLZ1, DTLZ2, DTLZ3 and DTLZ4.}                       
\label{PFsOfDTLZ}                                                        
\end{figure}

To calculate the IGD value of a result set $S$ of an MOEA running on an MOP, a set $R$ of representative points of the PF needs to be given in advance.
For DTLZ1 to DTLZ4,  we take the set of the intersecting points of weight vectors
and the PF surface as $R$.
Let $f^*=(f_{1}^*,...,f_{M}^*) $ be the intersecting point of a weight vector $w=(w_1,...,w_M)^T$ and the PF surface.
Then $f_i^*$ can be computed as \citep{MOEADD}
\begin{equation}
f_i^*=0.5\times\frac{w_i}{\sum_{j=1}^{M}w_j}
\end{equation}
for DTLZ1, and
\begin{equation}
f_i^*=\frac{w_i}{\sqrt{\sum_{j=1}^{M}w_j}}
\end{equation}
for DTLZ2, DTLZ3 and DTLZ4. The representative points for the global PFs of the 3-
objective instances of problems DTLZ1 to DTLZ4 are shown in Fig. \ref{PFsOfDTLZ}.

\subsection{WFG test suite}
The WFG test suite \citep{WFGProblems,WFG} allows test problem designers to construct scalable test problems with any number of objectives,
in which features such as modality and separability can be customized as required.
As discussed in \citep{WFGProblems,WFG},
it exceeds the functionality of the DTLZ test suite.
In particular, one can construct non-separable problems, deceptive problems,
truly degenerative problems, mixed shape PF problems,
problems scalable in the number of position-related parameters,
and problems with dependencies between position- and distance-related parameters as well with
the WFG test suite.

In \citep{WFG}, several scalable problems, i.e., WFG1 to WFG9,
are suggested for MOEA designers to test their algoritms,
which can be described as follows.
\begin{equation}
\begin{split}
Minimize \quad F(X)&=(f_1(X),...,f_M(X))\\
f_i(X)&=x_M+2ih_i(x_1,...,x_{M-1}) \\
X&=(x_1,...,x_M)^T
\end{split}
\end{equation}
where $h_i$ is a problem-dependent shape function  determining the geometry of the fitness
space, and $X$ is derived from a vector of working parameters $Z=(z_1,...,z_n)^T, z_i\in [0,2i]$ , by employing four problem-dependent transformation functions $t_1$, $t_2$, $t_3$  and  $t_4$.
Transformation functions must be designed carefully such that the underlying PF remains intact with a relatively easy to determine Pareto optimal set.
The WFG Toolkit provides a series of predefined shape and transformation functions to help ensure this is the case.
One can refer to  \citep{WFGProblems,WFG} to see their definitions.
Let
\begin{equation}
\begin{split}
Z''&=(z''_1,...,z''_m)^T
=t_4(t_3 (t_2 (t_1(Z'))))\\
Z'&=(z_1/2,...,z_n/2n)^T.
\end{split}
\end{equation}
Then $x_i=z''_i(z''_i-0.5)+0.5$ for problem WFG3, whereas $X=Z''$ for problems WFG1, WFG2 and WFG4 to WFG9.

The features of WFG1 to WFG9 can be summarized as follows.
\begin{itemize}
  \item WFG1:A separable and uni-modal problem with a biased PF and a convex and mixed geometry.
  \item WFG2:A non-separable problem with a convex and disconnected geometry, i.e., the PF of WFG2 is composed of several disconnected convex segments. And all of its objectives but $f_M$ are uni-modal.
  \item WFG3:A non-separable and uni-modal problem with a linear and degenerate PF shape, which can be seen as a connected version of WFG2.
  \item WFG4:A separable and multi-modal problem with large "hill sizes", and a concave geometry.
  \item WFG5:A separable and deceptive problem with a concave geometry.
  \item WFG6:A nonseparable and uni-modal problem with a concave geometry.
  \item WFG7:A separable and uni-modal problem with parameter dependency, and a concave geometry.
  \item WFG8:A nonseparable and uni-modal problem with parameter dependency, and a concave geometry.
  \item WFG9:A nonseparable, deceptive and uni-modal problem with parameter dependency, and a concave geometry.
\end{itemize}

As it can be seen from above, WFG1 and WFG7 are both separable and uni-modal,
and WFG8 and WFG9 have nonseparable property,
but the parameter dependency of WFG8 is much harder than that of WFG9.
In addition, the deceptiveness of WFG5 is more difficult than that of WFG9,
since WFG9 is only deceptive on its position parameters.
However, when it comes to the nonseparable reduction, WFG6 and WFG9 are more difficult than  WFG2 and WFG3.
Meanwhile, problems WFG4 to WFG9 share the same EF shape in the objective space,
which is a part of a hyper-ellipse with radii $r_i = 2i$, where $i\in\{1,...,M\}$.

\section{Experimental Results}
MOEA/DLD is implemented and run in the framework of jMetal5.4 \citep{jMetal2010,jMetal2011,jMetal2015}.
The data used for comparison  are all from paper \citep{MOEADD}.
Since MOEA/DD wins almost all the values on problems DTLZ1 to DTLZ4 in \citep{MOEADD}, we only keep its data for comparison. As for problems WFG1 to WFG9, GrEA performs the best in some values and the experimental data of it on problems WFG1 to WFG9  are also preserved.

\subsection{Parameter Settings}
The parameter settings of MOEA/DLD are listed as follows.
\begin{enumerate}
  \item Settings for Crossover Operator:The crossover probability is set as $p_c=1.0$ and the distribution index is $\eta_c=20$.
  \item Settings for Mutation Operator:The mutation probability is set as $p_m=1/n$. The distribution index is set as $\eta_m=20$.
  \item Population Size:The population size of MOEA/DLD is the same as the number of the weight vectors that can be calculated by Eq.(\ref{numberOfWeightVectors}). Since the divisions for 3- and 5-objective instances are set to 12 and 6, and the population sizes of them are 91 and 210, respectively. As for 8-, 10- and 15-objective instances, two-layer weight vector generation method is applied.  The divisions and the population sizes of them are listed in Table \ref{nDivisions}.
  \item Number of Runs:The algorithm is independently run 20 times on each test instance, which is the same as that of other algorithms for comparison.
  \item Number of Generations: All of the algorithms stopped at a predefined number of generations. The number of generations for DTLZ1 to DTLZ4 is listed in Table \ref{nGens}, and the number of generations for all the instances of WFG1 to WFG9 is 3000.
  \item Penalty Parameter in PBI: $ \theta= 5.0$.
  \item Neighborhood Size: $T = 20$.
  \item Selection Probability: The probability of selecting two mating individuals from the current neighborhood is set as $p_s = 0.8$.
  \item Settings for DTLZ1 to DTLZ4:As in papers \citep{IDBEA,MOEADD},
  the number of the objectives are set as $M \in \{3,5,8,10,15\}$ for comparative purpose.
  And the number of the decision variables is set as $n = M + r-1$, where $r = 5$ for DTLZ1, and $r = 10$ for DTLZ2, DTLZ3 and DTLZ4.
  To calculate the HV value  we set the reference point to $(1,...,1)^T$ for DTLZ1,  and $(2,...,2)^T$ for DTLZ2 to DTLZ4.
  \item Settings for WFG1 to WFG9:
  The number of the decision variables is set as n = k + l,
  where   $k = 2\times(M-1)$  is the position-related variable and $l = 20$ is the distance-related variable.
  To calculate the HV values for problems WFG1 to WFG9, the reference point is set to $(3,...,2M+1)^T$.
\end{enumerate}

\begin{table}
\caption{Population Sizes}
\begin{center}\label{nDivisions}
\begin{tabular}{|c|c|c|c|}
\hline
M&D1&D2&Population Size\\
\hline
3&12&-&91\\
5&6&-&210\\
8&3&2&156\\
10&3&2&275\\
15&2&1&135\\
\hline
\end{tabular}
\end{center}
\end{table}

\begin{table}
\caption{Number OF Generations}
\begin{center}
\begin{tabular}{|c|c|c|c|c|c|}

  \hline
  Instance& $M=3$ & $M=5$ & $M=8$ & $M=10$ & $M=15$ \\
  \hline
 DTLZ1 & 400 & 600 & 750 & 1000 & 1500 \\
  DTLZ2 & 250 & 350 & 500 & 750 & 1000 \\
  DTLZ3 & 1000 & 1000 & 1000 & 1500 & 2000 \\
  DTLZ4 & 600 & 1000 & 1250 & 2000 & 3000 \\
  \hline
\end{tabular}\label{nGens}
\end{center}
\end{table}

\subsection{Performance Comparisons on DTLZ1 to DTLZ4}

\begin{table*}[htbp]
  \centering
  \caption{Best, Median and Worst IGD Values by MOEA/DLD and MOEA/DD  on Instances of DTLZ1, DTLZ2, DTLZ3 and DTLZ4 with Different Number of Objectives.
The best performance is highlighted in gray background.}
  \resizebox{\linewidth}{!}{
    \begin{tabular}{|c|c|r|r|c|r|r|r|r|r|r|r|r|}
    \toprule
    Instance & M     & \multicolumn{1}{c|}{MOEA/DLD} & \multicolumn{1}{c|}{MOEA/DD} & Instance & \multicolumn{1}{c|}{MOEA/DLD} & \multicolumn{1}{c|}{MOEA/DD} & \multicolumn{1}{c|}{Instance} & \multicolumn{1}{c|}{MOEA/DLD} & \multicolumn{1}{c|}{MOEA/DD} & \multicolumn{1}{c|}{Instance} & \multicolumn{1}{c|}{MOEA/DLD} & \multicolumn{1}{c|}{MOEA/DD} \\
    \midrule
    \multirow{15}[10]{*}{DTLZ1} & \multirow{3}[2]{*}{3} & 4.893E-04 & \cellcolor[rgb]{ .851,  .851,  .851}3.191E-04 & \multirow{15}[10]{*}{DTLZ2} & 6.849E-04 & \cellcolor[rgb]{ .851,  .851,  .851}6.666E-04 & \multicolumn{1}{r|}{\multirow{15}[10]{*}{DTLZ3}} & \cellcolor[rgb]{ .851,  .851,  .851}3.228E-04 & 5.690E-04 & \multicolumn{1}{r|}{\multirow{15}[10]{*}{DTLZ4}} & 1.189E-04 & \cellcolor[rgb]{ .851,  .851,  .851}1.025E-04 \\
          &       & 1.052E-03 & \cellcolor[rgb]{ .851,  .851,  .851}5.848E-04 &       & \cellcolor[rgb]{ .851,  .851,  .851}7.623E-04 & 8.073E-04 &       & 2.098E-03 & \cellcolor[rgb]{ .851,  .851,  .851}1.892E-03 &       & 1.437E-04 & \cellcolor[rgb]{ .851,  .851,  .851}1.429E-04 \\
          &       & 4.106E-03 & \cellcolor[rgb]{ .851,  .851,  .851}6.573E-04 &       & \cellcolor[rgb]{ .851,  .851,  .851}1.023E-03 & 1.243E-03 &       & 7.162E-03 & \cellcolor[rgb]{ .851,  .851,  .851}6.231E-03 &       & \cellcolor[rgb]{ .851,  .851,  .851}1.717E-04 & 1.881E-04 \\
\cmidrule{2-4}\cmidrule{6-7}\cmidrule{9-10}\cmidrule{12-13}          & \multirow{3}[2]{*}{5} & 4.765E-04 & \cellcolor[rgb]{ .851,  .851,  .851}2.635E-04 &       & \cellcolor[rgb]{ .851,  .851,  .851}9.663E-04 & 1.128E-03 &       & \cellcolor[rgb]{ .851,  .851,  .851}4.670E-04 & 6.181E-04 &       & \cellcolor[rgb]{ .851,  .851,  .851}9.847E-05 & 1.097E-04 \\
          &       & 5.948E-04 & \cellcolor[rgb]{ .851,  .851,  .851}2.916E-04 &       & \cellcolor[rgb]{ .851,  .851,  .851}1.167E-03 & 1.291E-03 &       & \cellcolor[rgb]{ .851,  .851,  .851}8.756E-04 & 1.181E-03 &       & \cellcolor[rgb]{ .851,  .851,  .851}1.151E-04 & 1.296E-04 \\
          &       & 8.402E-04 & \cellcolor[rgb]{ .851,  .851,  .851}3.109E-04 &       & \cellcolor[rgb]{ .851,  .851,  .851}1.343E-03 & 1.424E-03 &       & \cellcolor[rgb]{ .851,  .851,  .851}2.168E-03 & 4.736E-03 &       & \cellcolor[rgb]{ .851,  .851,  .851}1.376E-04 & 1.532E-04 \\
\cmidrule{2-4}\cmidrule{6-7}\cmidrule{9-10}\cmidrule{12-13}          & \multirow{3}[2]{*}{8} & 2.846E-03 & \cellcolor[rgb]{ .851,  .851,  .851}1.809E-03 &       & \cellcolor[rgb]{ .851,  .851,  .851}2.327E-03 & 2.880E-03 &       & \cellcolor[rgb]{ .851,  .851,  .851}3.822E-03 & 3.411E-03 &       & 9.335E-04 & \cellcolor[rgb]{ .851,  .851,  .851}5.271E-04 \\
          &       & 3.830E-03 & \cellcolor[rgb]{ .851,  .851,  .851}2.589E-03 &       & \cellcolor[rgb]{ .851,  .851,  .851}3.200E-03 & 3.291E-03 &       & \cellcolor[rgb]{ .851,  .851,  .851}5.431E-03 & 8.079E-03 &       & 1.221E-03 & \cellcolor[rgb]{ .851,  .851,  .851}6.699E-04 \\
          &       & 5.067E-03 & \cellcolor[rgb]{ .851,  .851,  .851}2.996E-03 &       & \cellcolor[rgb]{ .851,  .851,  .851}3.827E-03 & 4.106E-03 &       & \cellcolor[rgb]{ .851,  .851,  .851}1.016E-02 & 1.826E-02 &       & 2.291E-01 & \cellcolor[rgb]{ .851,  .851,  .851}9.107E-04 \\
\cmidrule{2-4}\cmidrule{6-7}\cmidrule{9-10}\cmidrule{12-13}          & \multirow{3}[2]{*}{10} & 2.093E-03 & \cellcolor[rgb]{ .851,  .851,  .851}1.828E-03 &       & \cellcolor[rgb]{ .851,  .851,  .851}1.908E-03 & 3.223E-03 &       & \cellcolor[rgb]{ .851,  .851,  .851}1.600E-03 & 1.689E-03 &       & \cellcolor[rgb]{ .851,  .851,  .851}6.944E-04 & 1.291E-03 \\
          &       & 2.702E-03 & \cellcolor[rgb]{ .851,  .851,  .851}2.225E-03 &       & \cellcolor[rgb]{ .851,  .851,  .851}2.247E-03 & 3.752E-03 &       & \cellcolor[rgb]{ .851,  .851,  .851}1.982E-03 & 2.164E-03 &       & \cellcolor[rgb]{ .851,  .851,  .851}8.092E-04 & 1.615E-03 \\
          &       & 3.825E-03 & \cellcolor[rgb]{ .851,  .851,  .851}2.467E-03 &       & \cellcolor[rgb]{ .851,  .851,  .851}2.481E-03 & 4.145E-03 &       & \cellcolor[rgb]{ .851,  .851,  .851}2.863E-03 & 3.226E-03 &       & \cellcolor[rgb]{ .851,  .851,  .851}1.008E-03 & 1.931E-03 \\
\cmidrule{2-4}\cmidrule{6-7}\cmidrule{9-10}\cmidrule{12-13}          & \multirow{3}[2]{*}{15} & 4.083E-03 & \cellcolor[rgb]{ .851,  .851,  .851}2.867E-03 &       & 5.931E-03 & \cellcolor[rgb]{ .851,  .851,  .851}4.557E-03 &       & 6.306E-03 & \cellcolor[rgb]{ .851,  .851,  .851}5.716E-03 &       & \cellcolor[rgb]{ .851,  .851,  .851}1.082E-03 & 1.474E-03 \\
          &       & 4.925E-03 & \cellcolor[rgb]{ .851,  .851,  .851}4.203E-03 &       & 6.940E-03 & \cellcolor[rgb]{ .851,  .851,  .851}5.863E-03 &       & 7.763E-03 & \cellcolor[rgb]{ .851,  .851,  .851}7.461E-03 &       & \cellcolor[rgb]{ .851,  .851,  .851}1.697E-03 & 1.881E-03 \\
          &       & 8.312E-03 & \cellcolor[rgb]{ .851,  .851,  .851}4.669E-03 &       & 9.247E-03 & \cellcolor[rgb]{ .851,  .851,  .851}6.929E-03 &       & \cellcolor[rgb]{ .851,  .851,  .851}1.038E-02 & 1.138E-02 &       & 4.814E-03 & \cellcolor[rgb]{ .851,  .851,  .851}3.159E-03 \\
    \bottomrule
    \end{tabular}}%
  \label{IGDonDTLZs}%
\end{table*}%

We calculate the IGD values of the solution sets found by MOEA/DLD,
and compare the calculation results with those of MOEA/DD in \citep{MOEADD}.
As it is seen from Table \ref{IGDonDTLZs}, MOEA/DLD loses in all of the instances of DTLZ1, but shows competitive performance in the instances of DTLZ2 to DTLZ4. In detail, MOEA/DLD wins 11 out of 15 values for the instances of DTLZ2 and DTLZ3, and 9 out of 15 values for  the instances of DTLZ4.
In other words, MOEA/DD is a better optimizer for DTLZ1, but MOEA/DLD is a better optimizer for DTLZ2 to DTLZ4.
On the whole, MOEA/DLD shows competitive performance on DTLZ1 to DTLZ4 in terms of convergence and distribution, since the IGD values can characterize both convergence and distribution of the result sets.

\subsection{Performance Comparisons on WFG1 to WFG9}
The HV values of the result sets obtained by MOEA/DLD on WFG1 to WFG9 are calculated and compared to those of MOEA/DD and GrEA appeared in \citep{MOEADD}. All of the data is listed in Table \ref{HVonWFG}.

\begin{itemize}
  \item WFG1: MOEA/DD wins 11 of the 12 HV values, and hence is the best optimizer for WFG1.
  \item WFG2: MOEA/DLD wins 8 of the 12 HV values and MOEA/DD wins 4. It can be seen that MOEA/DLD is the best optimizer for the 3-objective instance, but it is hard to say that MOEA/DLD is superior to MOEA/DD or MOEA/DD is superior to MOEA/DLD for the 5-, 8- and 10-objective instances.
  \item WFG3: All of the three algorithms have their own advantages, and it is hard to say which one is the best for WFG3. But it can be seen that MOEA/DD is the best optimizer for the 3-objective instance, and GrEA is the best optimizer for the 8-objective instance.
  \item WFG4-WFG8: MOEA/DLD wins 54 of the 60 HV values of WFG4 to WFG8, and each of the rest 8 values is close to the best. Therefore, MOEA/DLD can be considered as the best optimizer for WFG4 to WFG8.
  \item WFG9: As it can be seen, MOEA/DD is the best optimizer for the 5-objective instance and GrEA is the best for the 8-objective instance. But it is hard to tell which one of the three algorithms is the best optimizer for the 3- and 10-objective instances.
\end{itemize}

On the whole, MOEA/DLD shows competitive performance on WFG1 to WFG9, especially on WFG4 to WFG8, in terms of convergence and diversity.
\begin{table*}[htbp]
  \centering
  \caption{Best, Median and Worst HV Values by MOEA/DLD, MOEA/DD and GrEA on instances of WFG1 to WFG9 with Different Number of Objectives. The best performance is highlighted in gray background.}
  \resizebox{\linewidth}{!}{
    \begin{tabular}{|c|c|c|c|c|c|c|c|c|c|c|c|c|}
    \toprule
    Instance & M     & MOEA/DLD & MOEA/DD & GrEA  & Instance & MOEA/DLD & MOEA/DD & GrEA  & Instance & MOEA/DLD & MOEA/DD & GrEA \\
    \midrule
    \multirow{12}[8]{*}{WFG1} & \multirow{3}[2]{*}{3} & 0.926005  & \cellcolor[rgb]{ .851,  .851,  .851}0.937694  & 0.794748  & \multirow{12}[8]{*}{WFG2} & \cellcolor[rgb]{ .851,  .851,  .851}0.959218  & 0.958287  & 0.950084  & \multirow{12}[8]{*}{WFG3} & 0.703568  & \cellcolor[rgb]{ .851,  .851,  .851}0.703664  & 0.699502  \\
          &       & 0.913039  & \cellcolor[rgb]{ .851,  .851,  .851}0.933402  & 0.692567  &       & \cellcolor[rgb]{ .851,  .851,  .851}0.958389  & 0.952467  & 0.942908  &       & 0.700921  & \cellcolor[rgb]{ .851,  .851,  .851}0.702964  & 0.672221  \\
          &       & \cellcolor[rgb]{ .851,  .851,  .851}0.904801  & 0.899253  & 0.627963  &       & \cellcolor[rgb]{ .851,  .851,  .851}0.811284  & 0.803397  & 0.800186  &       & 0.698947  & \cellcolor[rgb]{ .851,  .851,  .851}0.701624  & 0.662046  \\
\cmidrule{2-5}\cmidrule{7-9}\cmidrule{11-13}          & \multirow{3}[2]{*}{5} & 0.889460  & \cellcolor[rgb]{ .851,  .851,  .851}0.963464  & 0.876644  &       & \cellcolor[rgb]{ .851,  .851,  .851}0.995661  & 0.986572  & 0.980806  &       & 0.685894  & 0.673031  & \cellcolor[rgb]{ .851,  .851,  .851}0.695221  \\
          &       & 0.882779  & \cellcolor[rgb]{ .851,  .851,  .851}0.960897  & 0.831814  &       & \cellcolor[rgb]{ .851,  .851,  .851}0.994713  & 0.985129  & 0.976837  &       & 0.682373  & 0.668938  & \cellcolor[rgb]{ .851,  .851,  .851}0.684583  \\
          &       & 0.833721  & \cellcolor[rgb]{ .851,  .851,  .851}0.959840  & 0.790367  &       & 0.814734  & \cellcolor[rgb]{ .851,  .851,  .851}0.980035  & 0.808125  &       & \cellcolor[rgb]{ .851,  .851,  .851}0.677415  & 0.662951  & 0.671553  \\
\cmidrule{2-5}\cmidrule{7-9}\cmidrule{11-13}          & \multirow{3}[2]{*}{8} & 0.847126  & \cellcolor[rgb]{ .851,  .851,  .851}0.922284  & 0.811760  &       & \cellcolor[rgb]{ .851,  .851,  .851}0.983824  & 0.981673  & 0.980012  &       & 0.566758  & 0.598892  & \cellcolor[rgb]{ .851,  .851,  .851}0.657744  \\
          &       & 0.836809  & \cellcolor[rgb]{ .851,  .851,  .851}0.913024  & 0.681959  &       & 0.890574  & \cellcolor[rgb]{ .851,  .851,  .851}0.967265  & 0.840293  &       & 0.528185  & 0.565609  & \cellcolor[rgb]{ .851,  .851,  .851}0.649020  \\
          &       & 0.808549  & \cellcolor[rgb]{ .851,  .851,  .851}0.877784  & 0.616006  &       & 0.784746  & \cellcolor[rgb]{ .851,  .851,  .851}0.789739  & 0.778291  &       & 0.497817  & 0.556725  & \cellcolor[rgb]{ .851,  .851,  .851}0.638147  \\
\cmidrule{2-5}\cmidrule{7-9}\cmidrule{11-13}          & \multirow{3}[2]{*}{10} & 0.882825  & \cellcolor[rgb]{ .851,  .851,  .851}0.926815  & 0.866298  &       & \cellcolor[rgb]{ .851,  .851,  .851}0.987987  & 0.968201  & 0.964235  &       & \cellcolor[rgb]{ .851,  .851,  .851}0.563442  & 0.552713  & 0.543352  \\
          &       & 0.852469  & \cellcolor[rgb]{ .851,  .851,  .851}0.919789  & 0.832016  &       & \cellcolor[rgb]{ .851,  .851,  .851}0.982720  & 0.965345  & 0.959740  &       & \cellcolor[rgb]{ .851,  .851,  .851}0.533316  & 0.532897  & 0.513261  \\
          &       & 0.847490  & \cellcolor[rgb]{ .851,  .851,  .851}0.864689  & 0.757841  &       & 0.788291  & \cellcolor[rgb]{ .851,  .851,  .851}0.961400  & 0.956533  &       & 0.481038  & \cellcolor[rgb]{ .851,  .851,  .851}0.504943  & 0.501210  \\
    \midrule
    \multirow{12}[8]{*}{WFG4} & \multirow{3}[2]{*}{3} & \cellcolor[rgb]{ .851,  .851,  .851}0.731316  & 0.727060  & 0.723403  & \multirow{12}[8]{*}{WFG5} & \cellcolor[rgb]{ .851,  .851,  .851}0.698293  & 0.693665  & 0.689784  & \multirow{12}[8]{*}{WFG6} & \cellcolor[rgb]{ .851,  .851,  .851}0.711527  & 0.708910  & 0.699876  \\
          &       & \cellcolor[rgb]{ .851,  .851,  .851}0.730902  & 0.726927  & 0.722997  &       & 0.692802  & \cellcolor[rgb]{ .851,  .851,  .851}0.693544  & 0.689177  &       & \cellcolor[rgb]{ .851,  .851,  .851}0.701685  & 0.699663  & 0.693984  \\
          &       & \cellcolor[rgb]{ .851,  .851,  .851}0.729062  & 0.726700  & 0.722629  &       & 0.686016  & \cellcolor[rgb]{ .851,  .851,  .851}0.691173  & 0.688885  &       & \cellcolor[rgb]{ .851,  .851,  .851}0.698393  & 0.689125  & 0.685599  \\
\cmidrule{2-5}\cmidrule{7-9}\cmidrule{11-13}          & \multirow{3}[2]{*}{5} & \cellcolor[rgb]{ .851,  .851,  .851}0.884675  & 0.876181  & 0.881161  &       & \cellcolor[rgb]{ .851,  .851,  .851}0.844413  & 0.833159  & 0.836232  &       & 0.854733  & 0.850531  & \cellcolor[rgb]{ .851,  .851,  .851}0.855839  \\
          &       & \cellcolor[rgb]{ .851,  .851,  .851}0.883572  & 0.875836  & 0.879484  &       & \cellcolor[rgb]{ .851,  .851,  .851}0.841885  & 0.832710  & 0.834726  &       & \cellcolor[rgb]{ .851,  .851,  .851}0.849764  & 0.838329  & 0.847137  \\
          &       & \cellcolor[rgb]{ .851,  .851,  .851}0.882220  & 0.875517  & 0.877642  &       & \cellcolor[rgb]{ .851,  .851,  .851}0.838282  & 0.830367  & 0.832212  &       & \cellcolor[rgb]{ .851,  .851,  .851}0.844114  & 0.828315  & 0.840637  \\
\cmidrule{2-5}\cmidrule{7-9}\cmidrule{11-13}          & \multirow{3}[2]{*}{8} & \cellcolor[rgb]{ .851,  .851,  .851}0.944045  & 0.920869  & 0.787287  &       & \cellcolor[rgb]{ .851,  .851,  .851}0.895409  & 0.852838  & 0.838183  &       & \cellcolor[rgb]{ .851,  .851,  .851}0.914902  & 0.876310  & 0.912095  \\
          &       & \cellcolor[rgb]{ .851,  .851,  .851}0.941156  & 0.910146  & 0.784141  &       & \cellcolor[rgb]{ .851,  .851,  .851}0.892112  & 0.846736  & 0.641973  &       & \cellcolor[rgb]{ .851,  .851,  .851}0.904816  & 0.863087  & 0.902638  \\
          &       & \cellcolor[rgb]{ .851,  .851,  .851}0.936658  & 0.902710  & 0.679178  &       & \cellcolor[rgb]{ .851,  .851,  .851}0.887815  & 0.830338  & 0.571933  &       & \cellcolor[rgb]{ .851,  .851,  .851}0.890014  & 0.844535  & 0.885712  \\
\cmidrule{2-5}\cmidrule{7-9}\cmidrule{11-13}          & \multirow{3}[2]{*}{10} & \cellcolor[rgb]{ .851,  .851,  .851}0.975130  & 0.913018  & 0.896261  &       & \cellcolor[rgb]{ .851,  .851,  .851}0.919506  & 0.848321  & 0.791725  &       & 0.940347  & 0.884394  & \cellcolor[rgb]{ .851,  .851,  .851}0.943454  \\
          &       & \cellcolor[rgb]{ .851,  .851,  .851}0.969175  & 0.907040  & 0.843257  &       & \cellcolor[rgb]{ .851,  .851,  .851}0.918268  & 0.841118  & 0.725198  &       & 0.922168  & 0.859986  & \cellcolor[rgb]{ .851,  .851,  .851}0.927443  \\
          &       & \cellcolor[rgb]{ .851,  .851,  .851}0.964297  & 0.888885  & 0.840257  &       & \cellcolor[rgb]{ .851,  .851,  .851}0.913703  & 0.829547  & 0.685882  &       & \cellcolor[rgb]{ .851,  .851,  .851}0.916854  & 0.832299  & 0.884145  \\
    \midrule
    \multirow{12}[8]{*}{WFG7} & \multirow{3}[2]{*}{3} & \cellcolor[rgb]{ .851,  .851,  .851}0.731814  & 0.727069  & 0.723229  & \multirow{12}[8]{*}{WFG8} & \cellcolor[rgb]{ .851,  .851,  .851}0.677366  & 0.672022  & 0.671845  & \multirow{12}[8]{*}{WFG9} & 0.703985  & \cellcolor[rgb]{ .851,  .851,  .851}0.707269  & 0.702489  \\
          &       & \cellcolor[rgb]{ .851,  .851,  .851}0.731702  & 0.727012  & 0.722843  &       & \cellcolor[rgb]{ .851,  .851,  .851}0.676376  & 0.670558  & 0.669762  &       & 0.643070  & \cellcolor[rgb]{ .851,  .851,  .851}0.687401  & 0.638103  \\
          &       & \cellcolor[rgb]{ .851,  .851,  .851}0.731554  & 0.726907  & 0.722524  &       & \cellcolor[rgb]{ .851,  .851,  .851}0.670661  & 0.668593  & 0.667948  &       & \cellcolor[rgb]{ .851,  .851,  .851}0.642545  & 0.638194  & 0.636575  \\
\cmidrule{2-5}\cmidrule{7-9}\cmidrule{11-13}          & \multirow{3}[2]{*}{5} & \cellcolor[rgb]{ .851,  .851,  .851}0.888251  & 0.876409  & 0.884174  &       & 0.807753  & \cellcolor[rgb]{ .851,  .851,  .851}0.818663  & 0.797496  &       & 0.807766  & \cellcolor[rgb]{ .851,  .851,  .851}0.834616  & 0.823916  \\
          &       & \cellcolor[rgb]{ .851,  .851,  .851}0.888058  & 0.876297  & 0.883079  &       & \cellcolor[rgb]{ .851,  .851,  .851}0.807176  & 0.795215  & 0.792692  &       & 0.752847  & \cellcolor[rgb]{ .851,  .851,  .851}0.797185  & 0.753683  \\
          &       & \cellcolor[rgb]{ .851,  .851,  .851}0.887919  & 0.874909  & 0.881305  &       & \cellcolor[rgb]{ .851,  .851,  .851}0.805616  & 0.792900  & 0.790693  &       & 0.749143  & \cellcolor[rgb]{ .851,  .851,  .851}0.764723  & 0.747315  \\
\cmidrule{2-5}\cmidrule{7-9}\cmidrule{11-13}          & \multirow{3}[2]{*}{8} & \cellcolor[rgb]{ .851,  .851,  .851}0.950027  & 0.920763  & 0.918742  &       & \cellcolor[rgb]{ .851,  .851,  .851}0.897970  & 0.876929  & 0.803050  &       & 0.828326  & 0.772671  & \cellcolor[rgb]{ .851,  .851,  .851}0.842953  \\
          &       & \cellcolor[rgb]{ .851,  .851,  .851}0.949486  & 0.917584  & 0.910023  &       & \cellcolor[rgb]{ .851,  .851,  .851}0.850681  & 0.845975  & 0.799986  &       & 0.806266  & 0.759369  & \cellcolor[rgb]{ .851,  .851,  .851}0.831775  \\
          &       & \cellcolor[rgb]{ .851,  .851,  .851}0.949012  & 0.906219  & 0.901292  &       & \cellcolor[rgb]{ .851,  .851,  .851}0.828236  & 0.730348  & 0.775434  &       & 0.760654  & 0.689923  & \cellcolor[rgb]{ .851,  .851,  .851}0.765730  \\
\cmidrule{2-5}\cmidrule{7-9}\cmidrule{11-13}          & \multirow{3}[2]{*}{10} & \cellcolor[rgb]{ .851,  .851,  .851}0.977771  & 0.927666  & 0.937582  &       & \cellcolor[rgb]{ .851,  .851,  .851}0.960050  & 0.896317  & 0.841704  &       & 0.846621  & 0.717168  & \cellcolor[rgb]{ .851,  .851,  .851}0.860676  \\
          &       & \cellcolor[rgb]{ .851,  .851,  .851}0.977412  & 0.923441  & 0.902343  &       & \cellcolor[rgb]{ .851,  .851,  .851}0.912141  & 0.844036  & 0.838256  &       & \cellcolor[rgb]{ .851,  .851,  .851}0.838753  & 0.717081  & 0.706632  \\
          &       & \cellcolor[rgb]{ .851,  .851,  .851}0.977203  & 0.917141  & 0.901477  &       & \cellcolor[rgb]{ .851,  .851,  .851}0.892887  & 0.715250  & 0.830394  &       & \cellcolor[rgb]{ .851,  .851,  .851}0.759374  & 0.696061  & 0.686917  \\
    \bottomrule
    \end{tabular}}%
  \label{HVonWFG}%
\end{table*}%

\section{Handling Constraints}
\subsection{Modifications on the Comparison Procedure}
In this section, we extends MOEA/DLD (denoted as C-MOEA/DLD) to solve constraint MOPs of the following type:
\begin{equation}\label{C-MOP}
\begin{aligned}
\text{Minimize}\quad    &F(x)=(f_1(x),f_2(x),...,f_M(x))^T,\\
\text{subject to}\quad  &g_j(x)\geq 0, \quad j=1,2,...,J,\\
                        &h_k(x)=0,      \quad k=1,2,...,K,\\
                        &x^{(L)}_i\leq x_i \leq x^{(U)}_i, \quad i=1,2,...,n.
\end{aligned}
\end{equation}
As suggested in \citep{NSGAIII-PartII}, the constraint violation value of
a solution x (denoted as CV(x)) of Eq.(\ref{C-MOP}) can be calculated as
\begin{equation}\label{CV}
CV(x)=\sum^{J}_{j=1}\langle g_j(x)\rangle+\sum^{K}_{k=1}|h_k(x)|,
\end{equation}
where $\langle g_j(x)\rangle$  takes the absolute value of
$g_j(x)$  if $g_j(x)< 0$, and takes 0 otherwise.
A solution s1 is better than the other one s2, when $CV(s1)< CV(s2)$.
In this sense, a feasible solution is always better than an infeasible one, since the CV of a feasible solution is always 0 and the CV of an infeasible solution is always greater than 0.
However, when  both solutions are feasible, we have to take other measures to distinguish which solution is better than the other.

\begin{algorithm}
\caption{Constrained Version of the Comparison Procedure}
\label{modified-Criterion}
\begin{algorithmic}[1]
\REQUIRE two solutions s1 and s2, and the associated weight vector w.
\ENSURE  true if s1 is better than s2, or false otherwise.
\IF{s1 is feasible \AND s2 is feasible}
\RETURN $compare(s1,s2,w)$;
\ELSIF{s1 is feasible \AND s2 is infeasible}
\RETURN \TRUE;
\ELSIF{s1 is infeasible \AND s2 is infeasible}
\RETURN $(cv(s1)<=cv(s2)\ \AND\  compare(s1,s2,w))$;
\ELSE
\RETURN \FALSE;
\ENDIF
\end{algorithmic}
\end{algorithm}

In MOEA/DLD,  a hybrid method based on dominance and the PBI function is proposed to compare two given solutions s1 and s2, which can be implemented as a comparison procedure denoted as $compare(s1,s2,w)$. Solution s2 is replaced by solution s1 only when $compare(s1,s2,w)$ returns true. In C-MOEA/DLD, a direct modification to $compare(s1,s2,w)$ is made to handle constraint.
Besides this, there are no other changes to MOEA/DLD.
The modified version of $compare(s1,s2,w)$ is given in Algorithm \ref{modified-Criterion}, which is designed to deal with the following four situations that occurs in comparing two solutions s1 and s2 of a given constraint MOP.
\begin{enumerate}
\item Both solutions s1 and s2 are feasible. In this case the original comparison procedure $compare(s1,s2,w)$ is called by its modified version to judge which solution is better than the other one.
\item Solution s1 is feasible and solution s2 is infeasible. Then, the modified procedure returns true.
\item Both solutions s1 and s2 are infeasible. In this case, the CVs of the two solutions and $compare(s1,s2,w)$ are considered.  The modified procedure return true only when both $CV(s1)<=CV(s2)$ and the return value of $compare(s1,s2,w)$ are true.
\item Solution s1 is infeasible and solution s2 is feasble. Then, the modified procedure returns false.
\end{enumerate}

\subsection{Constraint Test Instances}
Four constraint problems suggested in \citep{NSGAIII-PartII}, i.e., C1-DTLZ1, C2-DTLZ2, C3-DTLZ1 and C3-DTLZ4 are chosen for our experimental studies, which are the same as those in \citep{MOEADD} for the convenience of comparison. And for each test instance, the objective functions, the number of objectives and decision variables are set the same as its unconstrained version. The features of the four problems can be briefly summarized as follows.
\begin{enumerate}
\item C1-DTLZ1: This constraint problem can be obtained by adding the following constraint to DTLZ1 and keeping its objective functions unchanged:
\begin{equation}
c(x)=1-\frac{f_M(x)}{0.6}-\sum^{M-1}_{i=1}\frac{f_i(x)}{0.5}\geq 0.
\end{equation}
This makes the global PF of C1-DTLZ1 the same as that of DTLZ1, but only a small region of the objective space close to the global PF is feasible.
\item C2-DTLZ2: Similarly, this constraint problem can be obtained by adding the following constraint to DTLZ2 and keeping its objective functions unchanged:
\begin{equation}\label{constraintOfC2DTLZ2}
\begin{aligned}
&c(x)=-min\left\{p(x),q(x)\right\}\geq 0,\\
&p(x)=\min^{M}_{i=1}\left((f_i(x)-1)^2+\sum^{M}_{j=1,j\neq i}f_j^2(x)-r^2 \right),\\
&q(x)=\sum^{M}_{i=1}\left(f_i(x)-1/\sqrt{M}\right)^2-r^2,
\end{aligned}
\end{equation}
where $r = 0.4$, for $m = 3$ and 0.5, otherwise. Since some parts of the objective space along the global PF of DTLZ2 are made infeasible by the constaint above, the global PF of C2-DTLZ2 is composed of several disconnected parts of the global pF of DTLZ2.
\item C3-DTLZ1: Again, the objective functions are the same as in DTLZ1 problem, but the following M linear constaints are added:
\begin{equation}
c_i(x)=\sum^{M}_{j=1,j\neq i}f_j(x)+\frac{f_i(x)}{0.5}-1 \geq 0,
\end{equation}
where $i\in \{1,2,...,M\}$. Some portions of the added constraint surfaces constitute the global PF of C3-DTLZ1.
\item C3-DTLZ4: Similarly, this constraint problem is obtained by adding M quadratic constraints as follows to DTLZ4 and keeping its objective functions unchanged:
\begin{equation}
c_i(x)=\frac{f_i^2(x)}{4}+\sum^{M}_{j=1,j\neq i}f_j^2(x)-1 \geq 0,
\end{equation}
where $i\in \{1,2,...,M\}$. And similarly, some portions of the added constraint surfaces constitute the global PF of C3-DTLZ4.
\end{enumerate}

\subsection{Performance Metric}
We use the IGD introduced in Section IV as our performance metric.
The IGD values of the running results of C-MOEA/DLD on the four constraint problems with 3,5,8,10 and 15 objectives are calculated, and compared with those of other MOEAs appeared in \citep{MOEADD}.
To calculate the IGD values, the set R of uniformly distributed representative points of the PF is necessary.
As discussed before, the set of the intersecting points of weight vectors and the PF
surface is taken as R for DTLZ1 to DTLZ4. The method of generating the set R is also presented in Section IV.
Since the PF of C1-DTLZ1 is the same as DTLZ1, the set R used for DTLZ1 is adopted for C1-DTLZ1.
As for C2-DTLZ2, the set R used for DTLZ2 is also adopted, but those points that violate the constraint introduced by Eq.(\ref{constraintOfC2DTLZ2}) are abandoned.
As for C3-DTLZ1 and C3-DTLZ4,  the intersecting points of weight vectors and the constraint surfaces are taken as the representative points of the PFs.
The intersenting point, denoted as $F^*=(f^*_1,f^*_2,...,f^*_M)^T$, of a weight vector $w=(w1,w2,...,w_M)^T$ and the constraint surface can be calculated as
\begin{equation}
f^*_i=\max_{j\in \{1,...,M\}}w_i\times t_j,
\end{equation}
where
\begin{equation}
t_j=\frac{1}{2\times w_j+ \sum^{M}_{k=1,k\neq j}w_k}
\end{equation}
for C3-DTLZ1, and
\begin{equation}
t_j=\frac{1}{\sqrt{\sum^{M}_{k=1,k\neq j}w_k^2+w_j^2/4}}
\end{equation}
for C3-DTLZ4.

\begin{figure}[!htbp]
\begin{center}                                                       
\subfigure[C1-DTLZ1]{                    
\includegraphics[scale=0.4]{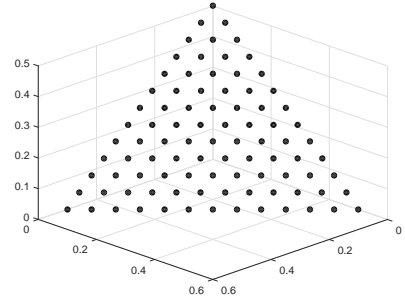}
}

\subfigure[C2-DTLZ2]{                    
\includegraphics[scale=0.4]{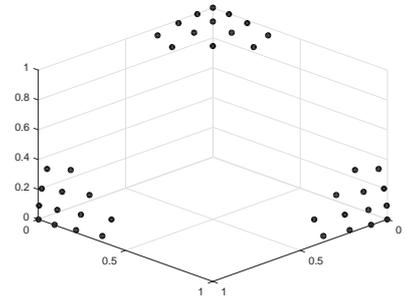}                
}

\subfigure[C3-DTLZ1]{                    
\includegraphics[scale=0.4]{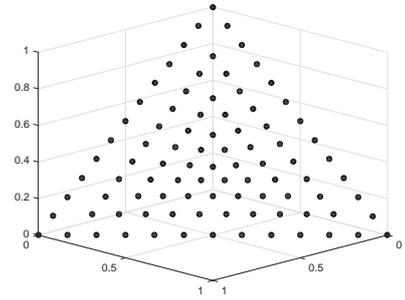}                
}

\subfigure[C3-DTLZ4]{                    
\includegraphics[scale=0.4]{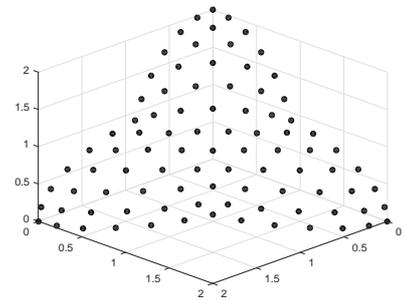}                
}
\end{center}

\caption{The representative points for the global PFs of the 3-objective instances of C1-DTLZ1, C2-DTLZ2, C3-DTLZ1 and C3-DTLZ4.}                       
\label{PFsOfConstraintProblems}                                                        
\end{figure}

The representative points for the global PFs of the 3-objective instances of the four constraint problems are shown in Fig. \ref{PFsOfConstraintProblems}.

\subsection{Performace comparisons}

\begin{table*}[htbp]
  \centering
  \caption{Best, Median and Worst IGD Values by C-MOEA/GLU and C-MOEA/DD on instances of C1-DTLZ1,C2-DTLZ2,C3-DTLZ1 and C3-DTLZ4  with Different Number of Objectives.  The best performance is highlighted in gray background.}
  \resizebox{\linewidth}{!}{
    \begin{tabular}{|c|c|c|c|c|c|c|c|c|c|c|c|c|}
    \toprule
    Instance & M     & CMOEA/DLD & CMOEA/DD & Instance & CMOEA/DLD & CMOEA/DD & Instance & CMOEA/DLD & CMOEA/DD & Instance & CMOEA/DLD & CMOEA/DD \\
    \midrule
    \multirow{15}[10]{*}{C1-DTLZ1} & \multirow{3}[2]{*}{3} & \cellcolor[rgb]{ .851,  .851,  .851}3.885E-04 & 4.239E-04 & \multirow{15}[10]{*}{C2-DTLZ2} & \cellcolor[rgb]{ .851,  .851,  .851}5.557E-04 & 6.623E-04 & \multirow{15}[10]{*}{C3-DTLZ1} & 1.498E-03 & \cellcolor[rgb]{ .851,  .851,  .851}1.018E-03 & \multirow{15}[10]{*}{C3-DTLZ4} & \cellcolor[rgb]{ .851,  .851,  .851}2.728E-03 & 3.944E-03 \\
          &       & 1.391E-03 & \cellcolor[rgb]{ .851,  .851,  .851}1.182E-03 &       & \cellcolor[rgb]{ .851,  .851,  .851}7.500E-04 & 8.708E-04 &       & \cellcolor[rgb]{ .851,  .851,  .851}2.927E-03 & 2.944E-03 &       & \cellcolor[rgb]{ .851,  .851,  .851}3.862E-03 & 4.968E-03 \\
          &       & 1.033E-02 & \cellcolor[rgb]{ .851,  .851,  .851}5.236E-03 &       & \cellcolor[rgb]{ .851,  .851,  .851}1.065E-03 & 6.688E-03 &       & 4.581E-02 & \cellcolor[rgb]{ .851,  .851,  .851}1.915E-02 &       & \cellcolor[rgb]{ .851,  .851,  .851}4.879E-03 & 6.716E-03 \\
\cmidrule{2-4}\cmidrule{6-7}\cmidrule{9-10}\cmidrule{12-13}          & \multirow{3}[2]{*}{5} & 6.829E-04 & \cellcolor[rgb]{ .851,  .851,  .851}6.091E-04 &       & \cellcolor[rgb]{ .851,  .851,  .851}9.574E-04 & 1.179E-03 &       & \cellcolor[rgb]{ .851,  .851,  .851}2.808E-04 & 4.647E-04 &       & \cellcolor[rgb]{ .851,  .851,  .851}2.214E-03 & 6.071E-03 \\
          &       & 1.505E-03 & \cellcolor[rgb]{ .851,  .851,  .851}1.939E-03 &       & \cellcolor[rgb]{ .851,  .851,  .851}1.196E-03 & 1.533E-03 &       & \cellcolor[rgb]{ .851,  .851,  .851}9.522E-04 & 1.847E-03 &       & \cellcolor[rgb]{ .851,  .851,  .851}2.720E-03 & 7.230E-03 \\
          &       & \cellcolor[rgb]{ .851,  .851,  .851}3.837E-03 & 9.282E-03 &       & \cellcolor[rgb]{ .851,  .851,  .851}1.386E-03 & 1.975E-03 &       & \cellcolor[rgb]{ .851,  .851,  .851}1.922E-03 & 2.163E-03 &       & \cellcolor[rgb]{ .851,  .851,  .851}3.415E-03 & 1.175E-02 \\
\cmidrule{2-4}\cmidrule{6-7}\cmidrule{9-10}\cmidrule{12-13}          & \multirow{3}[2]{*}{8} & 4.203E-03 & \cellcolor[rgb]{ .851,  .851,  .851}2.952E-03 &       & 6.028E-03 & \cellcolor[rgb]{ .851,  .851,  .851}1.875E-03 &       & \cellcolor[rgb]{ .851,  .851,  .851}1.521E-03 & 2.527E-03 &       & \cellcolor[rgb]{ .851,  .851,  .851}3.948E-03 & 1.116E-02 \\
          &       & \cellcolor[rgb]{ .851,  .851,  .851}5.358E-03 & 6.395E-03 &       & 7.692E-03 & \cellcolor[rgb]{ .851,  .851,  .851}3.258E-03 &       & \cellcolor[rgb]{ .851,  .851,  .851}5.119E-03 & 6.907E-03 &       & \cellcolor[rgb]{ .851,  .851,  .851}5.426E-03 & 1.386E-02 \\
          &       & 1.822E-02 & \cellcolor[rgb]{ .851,  .851,  .851}1.528E-02 &       & \cellcolor[rgb]{ .851,  .851,  .851}1.185E-02 & 8.997E-02 &       & \cellcolor[rgb]{ .851,  .851,  .851}7.118E-03 & 7.166E-03 &       & \cellcolor[rgb]{ .851,  .851,  .851}6.420E-03 & 1.974E-02 \\
\cmidrule{2-4}\cmidrule{6-7}\cmidrule{9-10}\cmidrule{12-13}          & \multirow{3}[2]{*}{10} & 3.156E-03 & \cellcolor[rgb]{ .851,  .851,  .851}2.564E-03 &       & 6.030E-03 & \cellcolor[rgb]{ .851,  .851,  .851}9.388E-04 &       & \cellcolor[rgb]{ .851,  .851,  .851}7.506E-04 & 2.698E-03 &       & \cellcolor[rgb]{ .851,  .851,  .851}3.406E-03 & 1.069E-02 \\
          &       & \cellcolor[rgb]{ .851,  .851,  .851}4.394E-03 & 4.511E-03 &       & 7.108E-03 & \cellcolor[rgb]{ .851,  .851,  .851}1.233E-03 &       & \cellcolor[rgb]{ .851,  .851,  .851}2.907E-03 & 4.646E-03 &       & \cellcolor[rgb]{ .851,  .851,  .851}3.884E-03 & 1.269E-02 \\
          &       & \cellcolor[rgb]{ .851,  .851,  .851}8.181E-03 & 1.004E-02 &       & 9.044E-03 & \cellcolor[rgb]{ .851,  .851,  .851}2.327E-03 &       & 5.139E-03 & \cellcolor[rgb]{ .851,  .851,  .851}4.841E-03 &       & \cellcolor[rgb]{ .851,  .851,  .851}4.994E-03 & 1.379E-02 \\
\cmidrule{2-4}\cmidrule{6-7}\cmidrule{9-10}\cmidrule{12-13}          & \multirow{3}[2]{*}{15} & 6.913E-03 & \cellcolor[rgb]{ .851,  .851,  .851}4.731E-03 &       & \cellcolor[rgb]{ .851,  .851,  .851}1.722E-03 & 2.589E-01 &       & \cellcolor[rgb]{ .851,  .851,  .851}2.219E-03 & 2.232E-03 &       & \cellcolor[rgb]{ .851,  .851,  .851}1.112E-02 & 2.212E-02 \\
          &       & \cellcolor[rgb]{ .851,  .851,  .851}1.310E-02 & 1.341E-02 &       & \cellcolor[rgb]{ .851,  .851,  .851}2.940E-03 & 3.996E-01 &       & 2.578E-03 & \cellcolor[rgb]{ .851,  .851,  .851}2.473E-03 &       & \cellcolor[rgb]{ .851,  .851,  .851}1.344E-02 & 2.602E-02 \\
          &       & \cellcolor[rgb]{ .851,  .851,  .851}2.747E-02 & 3.307E-02 &       & \cellcolor[rgb]{ .851,  .851,  .851}5.526E-03 & 1.063E+00 &       & 9.562E-03 & \cellcolor[rgb]{ .851,  .851,  .851}2.743E-03 &       & 9.557E-02 & \cellcolor[rgb]{ .851,  .851,  .851}3.171E-02 \\
    \bottomrule
    \end{tabular}}%
  \label{IGDonCDTLZ}%
\end{table*}%

The IGD values of the running results of C-MOEA/DLD on C1-DTLZ1, C2-DTLZ2, C3-DTLZ1 and C3-DTLZ4 are calculated and compared to those of C-MOEA/DD appeared in \citep{MOEADD}. As can be seen from Table \ref{IGDonCDTLZ}, C-MOEA/DLD shows competitive performance on the four constraint problems in terms of convergence and distribution.  More details are described as follows.
\begin{enumerate}
  \item C1-DTLZ1:MOEA/DLD wins 7 of the 15 IGD values, and MOEA/DD wins the rest 8 IGD values. It is hard to say which one of the two is a better optimizer for C1-DTLZ1.
  \item C2-DTLZ2:It can be seen that MOEA/DLD is the better optimizer for the 3-, 5- and 15-objective instances, and MOEA/DD is the better one for the 10-objective instance. But it is hard to say that MOEA/DLD is superior to MOEA/DD or the MOEA/DD is superior to MOEA/DLD for the 8-objective instance.
  \item C3-DTLZ1:MOEA/DLD is a better optimizer for the 5- and 8-objective instances, but  both of the algorithms have their own advantages in the rest 3 instances.
  \item C3-DTLZ4:MOEA/DLD is a better optimizer for the 3-, 5-, 8- and 10 instances, but outperformed by MOEA/DD on the worst IGD value of the 15-objective instance.
\end{enumerate}

\section{Conclusion}
In this paper, we propose an MOEA based on decomposition and local dominance for solving MaOPs, i.e., MOEA/DLD. The main ideas of MOEA/DLD can be summarized as follows. Firstly, MOEA/DLD employs a set of weight vectors to decompose a given MOP into a set of subproblems and optimizes them simultaneously, which is similar to other decomposition-based MOEAs.
Secondly, at each generation, the individuals in the old population and the new population are divided into $N$ subpopulations, each of which is associated with exactly one weight vector. The elitist strategy is then applied to keep $N$ elite individuals as the next generation.
Thirdly, each solution belongs to a subpopulation associated with exactly one weight vector.
The weight vector that has the smallest included angle with the solution is the associated weight vector of the solution.
Finally, in the elitist selection procedure, the domination method and the PBI function are only used to compare local solutions in each subpopulation.

Experimental results of MOEA/DLD on problems chosen from two famous test suites (i.e., DTLZ and WFG) with up to 15 objectives  are compared to those of other MOEAs appeared in \citep{MOEADD}.
Comparison results show that MOEA/DLD is very competitive in dealing with MaOPs , indicating that the proposed hybrid method based on decomposition and local dominance is effective.
We also extend MOEA/DLD to solve constraint problems. And the running results of the constrained version of MOEA/DLD (CMOEA/DLD) on C1-DTLZ1, C2-DTLZ2, C3-DTLZ1 and C3-DTLZ4 are compared to those of MOEA/DD. Comparison results shows that CMOEA/DLD has good performance in solving constraint MaOPs.

Our future work can be carried out in the following two aspects.
Firstly,  a recently proposed MaOP test suite \citep{MaOP} with complicated Pareto sets and uncommon PF shapes different from those of DTLZ and WFG will be used in our future works to observe  whether the performance of the proposed algorithm MOEA/DLD depends on the PF shapes. It has been reported in \citep{Hisao2017} that the performance of existing decomposition-based MOEAs is strongly dependent on the PF shapes. Therefore, if the performance of MODA/DLD also depends on the PF shapes, then it will be necessary to improve it to overcome this shortcoming.
Secondly, it is also interesting to study the performances of MOEA/DLD on other MOPs or constraint MOPs, such as the CEC2009 benchmark problems \citep{CEC2009}, the ZDT test problems \citep{ZDT},  combinatorial optimization problems \citep{Zitzler1999Multiobjective, Ishibuchi2010Many, Pan2015},
flowshop scheduling problems \citep{Pan2019, Ruiz2019213},
and especially some real-world problems with a large number of objectives.

\bibliographystyle{elsarticle-num}
\bibliography{moea}

\begin{thebibliography}{10}
\expandafter\ifx\csname url\endcsname\relax
  \def\url#1{\texttt{#1}}\fi
\expandafter\ifx\csname urlprefix\endcsname\relax\def\urlprefix{URL }\fi
\expandafter\ifx\csname href\endcsname\relax
  \def\href#1#2{#2} \def\path#1{#1}\fi

\bibitem{Panda2009937}
S.~Panda, Multi-objective evolutionary algorithm for sssc-based controller
  design, Electric Power Systems Research 79~(6) (2009) 937 -- 944.
\newblock \href
  {http://dx.doi.org/http://dx.doi.org/10.1016/j.epsr.2008.12.004}
  {\path{doi:http://dx.doi.org/10.1016/j.epsr.2008.12.004}}.

\bibitem{WBS2013}
G.~Fu, Z.~Kapelan, J.~R. Kasprzyk, P.~Reed, Optimal design of water
  distribution systems using many-objective visual analytics, Journal of Water
  Resources Planning \& Management 139~(6) (2013) 624--633.

\bibitem{AEC2013}
R.~J. Lygoe, M.~Cary, P.~J. Fleming, A real-world application of a
  many-objective optimisation complexity reduction process, in: International
  Conference on Evolutionary Multi-Criterion Optimization, 2013, pp. 641--655.

\bibitem{LUM2012}
O.~Chikumbo, E.~Goodman, K.~Deb, Approximating a multi-dimensional pareto front
  for a land use management problem: A modified moea with an epigenetic
  silencing metaphor, in: Evolutionary Computation, 2012, pp. 1--9.

\bibitem{Ganesan2015293}
T.~Ganesan, I.~Elamvazuthi, P.~Vasant, Multiobjective design optimization of a
  nano-cmos voltage-controlled oscillator using game theoretic-differential
  evolution, Applied Soft Computing 32 (2015) 293 -- 299.
\newblock \href
  {http://dx.doi.org/http://dx.doi.org/10.1016/j.asoc.2015.03.016}
  {\path{doi:http://dx.doi.org/10.1016/j.asoc.2015.03.016}}.

\bibitem{Ganesan2013}
T.~Ganesan, I.~Elamvazuthi, K.~Z.~K. Shaari, P.~Vasant, Hypervolume-Driven
  Analytical Programming for Solar-Powered Irrigation System Optimization,
  Springer International Publishing, Heidelberg, 2013, pp. 147--154.
\newblock \href {http://dx.doi.org/10.1007/978-3-319-00542-3-15}
  {\path{doi:10.1007/978-3-319-00542-3-15}}.

\bibitem{DomingoPerez201695}
F.~Domingo-Perez, J.~L. Lazaro-Galilea, A.~Wieser, E.~Martin-Gorostiza,
  D.~Salido-Monzu, A.~de~la Llana, Sensor placement determination for
  range-difference positioning using evolutionary multi-objective optimization,
  Expert Systems with Applications 47 (2016) 95 -- 105.
\newblock \href
  {http://dx.doi.org/http://dx.doi.org/10.1016/j.eswa.2015.11.008}
  {\path{doi:http://dx.doi.org/10.1016/j.eswa.2015.11.008}}.

\bibitem{Pan2011}
Q.-K. Pan, L.~Wang, L.~Gao, W.~Li,
  \href{http://www.sciencedirect.com/science/article/pii/S0020025510005086}{An
  effective hybrid discrete differential evolution algorithm for the flow shop
  scheduling with intermediate buffers}, Information Sciences 181~(3) (2011)
  668 -- 685.
\newblock \href {http://dx.doi.org/https://doi.org/10.1016/j.ins.2010.10.009}
  {\path{doi:https://doi.org/10.1016/j.ins.2010.10.009}}.
\newline\urlprefix\url{http://www.sciencedirect.com/science/article/pii/S0020025510005086}

\bibitem{Pan2018}
T.~Meng, Q.-K. Pan, J.-Q. Li, H.-Y. Sang,
  \href{http://www.sciencedirect.com/science/article/pii/S2210650216304965}{An
  improved migrating birds optimization for an integrated lot-streaming flow
  shop scheduling problem}, Swarm and Evolutionary Computation 38 (2018) 64 --
  78.
\newblock \href {http://dx.doi.org/https://doi.org/10.1016/j.swevo.2017.06.003}
  {\path{doi:https://doi.org/10.1016/j.swevo.2017.06.003}}.
\newline\urlprefix\url{http://www.sciencedirect.com/science/article/pii/S2210650216304965}

\bibitem{Pan2019}
Q.-K. Pan, L.~Gao, L.~Wang, J.~Liang, X.-Y. Li, Effective heuristics and
  metaheuristics to minimize total flowtime for the distributed permutation
  flowshop problem, Expert systems with Application.

\bibitem{Najafi201446}
B.~Najafi, A.~Shirazi, M.~Aminyavari, F.~Rinaldi, R.~A. Taylor, Exergetic,
  economic and environmental analyses and multi-objective optimization of an
  sofc-gas turbine hybrid cycle coupled with an \{MSF\} desalination system,
  Desalination 334~(1) (2014) 46 -- 59.
\newblock \href
  {http://dx.doi.org/http://dx.doi.org/10.1016/j.desal.2013.11.039}
  {\path{doi:http://dx.doi.org/10.1016/j.desal.2013.11.039}}.

\bibitem{PF}
K.~Miettinen, Nonlinear Multiobjective Optimization, Norwell, MA:Kluwer, 1999.

\bibitem{NSGAII}
K.~Deb, A.~Pratap, S.~Agarwal, T.~Meyarivan, A fast and elitist multiobjective
  genetic algorithm: Nsga-ii, IEEE Transactions on Evolutionary Computation
  6~(2) (2002) 182--197.
\newblock \href {http://dx.doi.org/10.1109/4235.996017}
  {\path{doi:10.1109/4235.996017}}.

\bibitem{SPEA2}
E.~Zitzler, M.~Laumanns, L.~Thiele, Spea2: Improving the strength pareto
  evolutionary algorithm for multiobjective optimization, in: Evolutionary
  Methods for Design, Optimisation, and Control, CIMNE, Barcelona, Spain, 2002,
  pp. 95--100.

\bibitem{MOEAD}
Q.~Zhang, H.~Li, Moea/d: A multiobjective evolutionary algorithm based on
  decomposition, IEEE Transactions on Evolutionary Computation 11~(6) (2007)
  712--731.
\newblock \href {http://dx.doi.org/10.1109/TEVC.2007.892759}
  {\path{doi:10.1109/TEVC.2007.892759}}.

\bibitem{Survey2017}
A.~Trivedi, D.~Srinivasan, K.~Sanyal, A.~Ghosh, A survey of multiobjective
  evolutionary algorithms based on decomposition, IEEE Transactions on
  Evolutionary Computation 21~(3) (2017) 440--462.
\newblock \href {http://dx.doi.org/10.1109/TEVC.2016.2608507}
  {\path{doi:10.1109/TEVC.2016.2608507}}.

\bibitem{Emmerich2005}
M.~Emmerich, N.~Beume, B.~Naujoks, An EMO Algorithm Using the Hypervolume
  Measure as Selection Criterion, Springer Berlin Heidelberg, Berlin,
  Heidelberg, 2005, pp. 62--76.
\newblock \href {http://dx.doi.org/10.1007/978-3-540-31880-4-5}
  {\path{doi:10.1007/978-3-540-31880-4-5}}.

\bibitem{NSGAIII}
K.~Deb, H.~Jain, An evolutionary many-objective optimization algorithm using
  reference-point-based nondominated sorting approach, part i: Solving problems
  with box constraints, IEEE Transactions on Evolutionary Computation 18~(4)
  (2014) 577--601.
\newblock \href {http://dx.doi.org/10.1109/TEVC.2013.2281535}
  {\path{doi:10.1109/TEVC.2013.2281535}}.

\bibitem{RVEA}
R.~Cheng, Y.~Jin, M.~Olhofer, B.~Sendhoff, A reference vector guided
  evolutionary algorithm for many-objective optimization, IEEE Transactions on
  Evolutionary Computation 20~(5) (2016) 773--791.
\newblock \href {http://dx.doi.org/10.1109/TEVC.2016.2519378}
  {\path{doi:10.1109/TEVC.2016.2519378}}.

\bibitem{MOEADD}
K.~Li, K.~Deb, Q.~Zhang, S.~Kwong, An evolutionary many-objective optimization
  algorithm based on dominance and decomposition, IEEE Transactions on
  Evolutionary Computation 19~(5) (2015) 694--716.
\newblock \href {http://dx.doi.org/10.1109/TEVC.2014.2373386}
  {\path{doi:10.1109/TEVC.2014.2373386}}.

\bibitem{SBX}
K.~Deb, R.~B. Agrawal, Simulated binary crossover for continuous search space,
  Complex Systems 9~(3) (2000) 115--148.

\bibitem{PM}
K.~Deb, M.~Goyal, A combined genetic adaptive search (geneas) for engineering
  design, 1996, pp. 30--45.

\bibitem{SSA}
I.~Das, J.~Dennis,
  \href{https://doi.org/10.1137/S1052623496307510}{Normal-boundary
  intersection: A new method for generating the pareto surface in nonlinear
  multicriteria optimization problems}, SIAM Journal on Optimization 8~(3)
  (1998) 631--657.
\newblock \href
  {http://arxiv.org/abs/https://doi.org/10.1137/S1052623496307510}
  {\path{arXiv:https://doi.org/10.1137/S1052623496307510}}, \href
  {http://dx.doi.org/10.1137/S1052623496307510}
  {\path{doi:10.1137/S1052623496307510}}.
\newline\urlprefix\url{https://doi.org/10.1137/S1052623496307510}

\bibitem{Cvetkovic2002MOP}
D.~Cvetkovic, I.~C. Parmee, Preferences and their application in evolutionary
  multiobjective optimization, Evolutionary Computation IEEE Transactions on
  6~(1) (2002) 42--57.

\bibitem{Osiadacz1989MOP}
A.~J. Osiadacz, Multiple criteria optimization; theory, computation, and
  application, Optimal Control Applications \& Methods 10~(1) (1989) 89¨C90.

\bibitem{Coit1998Genetic}
D.~Coit, Genetic algorithms and engineering design, Engineering Economist
  43~(4) (1998) 379--381.

\bibitem{IGD}
P.~A.~N. Bosman, D.~Thierens, The balance between proximity and diversity in
  multiobjective evolutionary algorithms, IEEE Transactions on Evolutionary
  Computation 7~(2) (2003) 174--188.
\newblock \href {http://dx.doi.org/10.1109/TEVC.2003.810761}
  {\path{doi:10.1109/TEVC.2003.810761}}.

\bibitem{HV}
E.~Zitzler, L.~Thiele, Multiobjective evolutionary algorithms: a comparative
  case study and the strength pareto approach, IEEE Transactions on
  Evolutionary Computation 3~(4) (1999) 257--271.
\newblock \href {http://dx.doi.org/10.1109/4235.797969}
  {\path{doi:10.1109/4235.797969}}.

\bibitem{HYPE}
J.~Bader, E.~Zitzler, Hype: An algorithm for fast hypervolume-based
  many-objective optimization, Evolutionary Computation 19~(1) (2011) 45--76.
\newblock \href {http://dx.doi.org/10.1162/EVCO\_a\_00009}
  {\path{doi:10.1162/EVCO\_a\_00009}}.

\bibitem{WFGalgorithm}
L.~While, L.~Bradstreet, L.~Barone, A fast way of calculating exact
  hypervolumes, IEEE Transactions on Evolutionary Computation 16~(1) (2012)
  86--95.
\newblock \href {http://dx.doi.org/10.1109/TEVC.2010.2077298}
  {\path{doi:10.1109/TEVC.2010.2077298}}.

\bibitem{DTLZ}
K.~Deb, L.~Thiele, M.~Laumanns, E.~Zitzler, Scalable test problems for
  evolutionary multiobjective optimization, Evolutionary Multiobjective
  Optimization (2001) 105--145.

\bibitem{WFGProblems}
S.~Huband, L.~Barone, L.~While, P.~Hingston, A scalable multi-objective test
  problem toolkit, Lecture Notes in Computer Science 3410 (2005) 280--295.

\bibitem{WFG}
S.~Huband, P.~Hingston, L.~Barone, L.~While, A review of multiobjective test
  problems and a scalable test problem toolkit, IEEE Transactions on
  Evolutionary Computation 10~(5) (2006) 477--506.
\newblock \href {http://dx.doi.org/10.1109/TEVC.2005.861417}
  {\path{doi:10.1109/TEVC.2005.861417}}.

\bibitem{jMetal2010}
J.~Durillo, A.~Nebro, E.~Alba, The jmetal framework for multi-objective
  optimization: Design and architecture, in: CEC 2010, Barcelona, Spain, 2010,
  pp. 4138--4325.

\bibitem{jMetal2011}
J.~J. Durillo, A.~J. Nebro,
  \href{http://www.sciencedirect.com/science/article/pii/S0965997811001219}{jmetal:
  A java framework for multi-objective optimization}, Advances in Engineering
  Software 42 (2011) 760--771.
\newblock \href {http://dx.doi.org/DOI: 10.1016/j.advengsoft.2011.05.014}
  {\path{doi:DOI: 10.1016/j.advengsoft.2011.05.014}}.
\newline\urlprefix\url{http://www.sciencedirect.com/science/article/pii/S0965997811001219}

\bibitem{jMetal2015}
A.~J. Nebro, J.~J. Durillo, M.~Vergne,
  \href{http://doi.acm.org/10.1145/2739482.2768462}{Redesigning the jmetal
  multi-objective optimization framework}, in: Proceedings of the Companion
  Publication of the 2015 Annual Conference on Genetic and Evolutionary
  Computation, GECCO Companion '15, ACM, New York, NY, USA, 2015, pp.
  1093--1100.
\newblock \href {http://dx.doi.org/10.1145/2739482.2768462}
  {\path{doi:10.1145/2739482.2768462}}.
\newline\urlprefix\url{http://doi.acm.org/10.1145/2739482.2768462}

\bibitem{IDBEA}
M.~Asafuddoula, T.~Ray, R.~Sarker, A decomposition-based evolutionary algorithm
  for many objective optimization, IEEE Transactions on Evolutionary
  Computation 19~(3) (2015) 445--460.
\newblock \href {http://dx.doi.org/10.1109/TEVC.2014.2339823}
  {\path{doi:10.1109/TEVC.2014.2339823}}.

\bibitem{NSGAIII-PartII}
H.~Jain, K.~Deb, An evolutionary many-objective optimization algorithm using
  reference-point based nondominated sorting approach, part ii: Handling
  constraints and extending to an adaptive approach, IEEE Transactions on
  Evolutionary Computation 18~(4) (2014) 602--622.
\newblock \href {http://dx.doi.org/10.1109/TEVC.2013.2281534}
  {\path{doi:10.1109/TEVC.2013.2281534}}.

\bibitem{MaOP}
H.~Li, K.~Deb, Q.~Zhang, P.~N. Suganthan, Challenging novel many and
  multi-objective bound constrained benchmark problems, Tech. rep. (2017).

\bibitem{Hisao2017}
H.~Ishibuchi, Y.~Setoguchi, H.~Masuda, Y.~Nojima, Performance of
  decomposition-based many-objective algorithms strongly depends on pareto
  front shapes, IEEE Transactions on Evolutionary Computation 21~(2) (2017)
  169--190.
\newblock \href {http://dx.doi.org/10.1109/TEVC.2016.2587749}
  {\path{doi:10.1109/TEVC.2016.2587749}}.

\bibitem{CEC2009}
Q.~Zhang, A.~Zhou, S.~Z. Zhao, P.~N. Suganthan, W.~Liu, S.~Tiwari,
  Multiobjective optimization test instances for the cec 2009 special session
  and competition, Tech. rep. (2008).

\bibitem{ZDT}
E.~Zitzler, K.~Deb, L.~Thiele,
  \href{https://doi.org/10.1162/106365600568202}{Comparison of multiobjective
  evolutionary algorithms: Empirical results}, Evolutionary Computation 8~(2)
  (2000) 173--195.
\newblock \href {http://arxiv.org/abs/https://doi.org/10.1162/106365600568202}
  {\path{arXiv:https://doi.org/10.1162/106365600568202}}, \href
  {http://dx.doi.org/10.1162/106365600568202}
  {\path{doi:10.1162/106365600568202}}.
\newline\urlprefix\url{https://doi.org/10.1162/106365600568202}

\bibitem{Zitzler1999Multiobjective}
E.~Zitzler, L.~Thiele, Multiobjective evolutionary algorithms: a comparative
  case study and the strength pareto approach, IEEE Transactions on
  Evolutionary Computation 3~(4) (1999) 257--271.

\bibitem{Ishibuchi2010Many}
H.~Ishibuchi, Y.~Hitotsuyanagi, N.~Tsukamoto, Y.~Nojima, Many-objective test
  problems to visually examine the behavior of multiobjective evolution in a
  decision space, in: International Conference on Parallel Problem Solving From
  Nature, 2010, pp. 91--100.

\bibitem{Pan2015}
B.~Zhang, Q.-K. Pan, X.-L. Zhang, P.-Y. Duan,
  \href{http://www.sciencedirect.com/science/article/pii/S1568494615000265}{An
  effective hybrid harmony search-based algorithm for solving multidimensional
  knapsack problems}, Applied Soft Computing 29 (2015) 288 -- 297.
\newblock \href {http://dx.doi.org/https://doi.org/10.1016/j.asoc.2015.01.022}
  {\path{doi:https://doi.org/10.1016/j.asoc.2015.01.022}}.
\newline\urlprefix\url{http://www.sciencedirect.com/science/article/pii/S1568494615000265}

\bibitem{Ruiz2019213}
R.~Ruiz, Q.-K. Pan, B.~Naderi,
  \href{http://www.sciencedirect.com/science/article/pii/S0305048317306990}{Iterated
  greedy methods for the distributed permutation flowshop scheduling problem},
  OMEGA-International Journal of Management Science 83 (2019) 213 -- 222.
\newblock \href {http://dx.doi.org/https://doi.org/10.1016/j.omega.2018.03.004}
  {\path{doi:https://doi.org/10.1016/j.omega.2018.03.004}}.
\newline\urlprefix\url{http://www.sciencedirect.com/science/article/pii/S0305048317306990}

\end{thebibliography}

\end{document}